# Graph Neural Networks: a bibliometrics overview

**Abdalsamad Keramatfar [1,2], Mohadeseh Rafiee[1], Hossein Amirkhani[2]**

[1] SID, Academic Center for Education, Culture and Research (ACECR) , Tehran, Iran
[2] Department of Computer Engineering and IT, Faculty of Engineering, University of Qom, Qom, Iran
Corresponding author: A. Keramatfar (e-mail: samad@sid.com).

## Abstract

Recently, graph neural networks (GNN) have become a hot topic in machine learning community. This paper presents a Scopus-based bibliometric overview of the GNNs' research since 2004, when GNN papers were first published. The study aims to evaluate GNN research trend, both quantitatively and qualitatively. We provide the trend of research, distribution of subjects, active and influential authors and institutions, sources of publications, most cited documents, and hot topics. Our investigations reveal that the most frequent subject categories in this field are computer science, engineering, telecommunications, linguistics, operations research and management science, information science and library science, business and economics, automation and control systems, robotics, and social sciences. In addition, the most active source of GNN publications is Lecture Notes in Computer Science. The most prolific or impactful institutions are found in the United States, China, and Canada. We also provide must-read papers and future directions. Finally, the application of graph convolutional networks and attention mechanism are now among hot topics of GNN research.

## 1. Introduction

Graph data is ubiquitous; world wide web, social networks including online social networks, biological and chemical structures including brain and protein networks, scientific networks including citations, collaborations, and co-occurrences networks, business networks including network of financial transactions, wireless sensor networks, knowledge graphs, and water networks are just a few examples of the real-world data which are inherently graphs. The wide range of graph data has grabbed wide attention from scientific community [1-6] and specially from data mining and machine learning communities [7] for fraud detection [8], sentiment analysis [9-11], link prediction [12], traffic forecasting [13], molecular graph generation [14], recommender systems [15, 16], epidemiology prediction [17], and action recognition [18-20].

Recently, deep learning [21] algorithms have achieved notable breakthrough in different areas including natural language processing [22], computer vision [23], and speech recognition [24]. The ability to extract high quality features by passing data through different non-linear layers is the main reason for this performance. Moreover, deep learning has provided special algorithms for conventional types of data. Recurrent neural networks (RNN) and their variants, and recently transformers [25], were presented for sequential data, like texts, and convolutional neural networks (CNN) were presented for machine vision. Deep learning community has mainly focused on 1D, 2D, and 3D Euclidean structured data, such as images, videos, and texts [26].

As mentioned earlier, there are many data types, which inherently resemble graphs. This motivated machine learning community to provide different algorithms for graph data including label propagation [27] and Laplacian regularization [28]. However, with the success of deep learning methods, there has recently been growing attention to geometric deep, a kind of learning attempting to generalize deep learning to non-Euclidean data, such as graphs and manifolds [29].

Due to their increasing popularity, deep learning-based methods have been developed for handling different graph tasks. Graph Neural Networks (GNNs) are deep learning-based models which work on graph data and have been widely utilized recently [30]. A successful subcategory of GNNs is motivated by CNNs which are state-of-the-art models on a variety of machine vision tasks [29]. CNNs have the ability to extract multi-scale spatial features using their filters to combine these features to construct high-quality representations [31]. However, it is hard for CNNs to mine and learn graph data [32]. As the structure of regular Euclidean data is a special kind of graph, machine learning community has tried to generalize the power of CNNs to graphs, too [30].

Generally, GNNs represent each node as a weighted sum of their neighborhood representations. If the signal is smooth enough with respect to the underlying graph, this feature aggregation process will provide extra information and, consequently better representation for the nodes. GNNs are connectionist models which hold states that can aggregate information from their neighbors with arbitrary distances. They model the graphs dependence via message-passing between nodes [30] and have been used in different graph-based tasks including node classification, graph classification, link prediction, and clustering [30, 33].

In the node classification problem, GNNs map network data to outputs in two steps. First, an information propagation occurs which provides node representations. These representations are in a low-dimensional space where neighboring nodes have similar representations [34]. Second, the model maps node representations to node labels [35]. For example, there are social theories which indicating that friends on social networks tend to express similar sentiments towards specific topics and a person tends to post tweets with similar sentiments [36]. These theories provide us with graphs in which social media posts constitute nodes, and different kinds of relations constitute edges. So, the extra textual features which exist in the neighborhood of a post can enrich its representation for node sentiment classification. In the graph-level classification problems, similar to CNN's terminology, the graph pooling/readout aims to obtain representation of the entire graph [34]. In the link prediction setting, a common approach is to first compute node representations by GNNs, and then combine the representations of two nodes as their interlink representation [37]. For the clustering problems, algorithms usually embed the nodes using a GNN. The node embeddings are then fed to a conventional unsupervised algorithm to identify the clusters [38].

In the last few years, GNNs have become a hot topic in machine learning and benefited many graph-related tasks [39, 40] and real world applications across a verity of areas, including molecule design [14], financial fraud detection [8], traffic prediction [8, 13], user behavior analysis [15, 41, 42], and recommender systems [16, 43]. In addition, GNNs have performed well in learning from non-structural data such as texts and images. In the NLP and text analysis field, GNN-based modeling has been used for different tasks, such as semantic role labeling [44], relation extraction [45], gender detection [46], reading comprehension [47], sentence generation [48], and sentiment analysis [49]. In the machine vision field, researchers have carried out analyses based on different types of graphs such as the pixel adjacency network in different tasks including hyperspectral image classification [50], action recognition [51], scene graph generation [52] , and multi-label image recognition [53].

Bibliometrics is a field of study which uses publications' bibliographic data and citation relations to evaluate and reveal the structure of a research field [54-58]. Previous research used bibliometric methods to analyze different subfields of computer science, such as sentiment analysis [59-61], granular computing [62], topic modelling , NLP [63-65], and deep learning [66, 67]. To the best of our knowledge, there are not any bibliometric studies [68], which target the young and fast-growing research field of GNNs. So, in this paper, we employ different bibliometric methods to shed new light on the patterns in the global GNNs research. Using these methods, we identify characteristics of research, top subject categories, researchers, institutions, papers, journals, and hot topics. The rest of this article is organized as follows: in Section 2, we introduce the setup of our study; Section 3 provides the results of the study, and we conclude our results in Section 4.

## 2. Material and methods

Scopus[1] is the largest abstract and indexing database of scientific publications and one of the main sources for bibliometrics studies [69-72]. Therefore, we used this database as our main source of data. In order to extract documents related to GNN, we used the following query in August 6, 2020:

TITLE-ABS-KEY= "graph neural network" OR "convolutional graph neural network" OR "graph convolutional network" OR ("graph representation learning" and (deep or neural or convolution)) OR "graph autoencoder" OR "geometric deep learning". TITLE-ABS-KEY is a Scopus reserved key which searches the title, abstract, and keywords of documents for the provided term. The study period spanned from 1960 (our earliest access to the Scopus records) to 2020.

We used the Bibexcel version 2008-08 [73] to discover the most common topics (identified by extracting the most frequent keywords) and subject categories, the most prolific authors, and h-index of researchers, the most prolific journals, and must-read papers. In addition, to assess the sources of publications qualitatively, we used SJR (SCImago Journal and Country Rank) of all sources from SCImago[2], which is a free database of journals and countries bibliometric indicators based on Scopus data. SJR is a journal bibliometric index similar to impact factor that weighs citations based on the reputation of the source journals [74, 75]. In other words, citations from more impactful journals are not considered equal to those from less impactful ones. To extract the most prolific and impactful institutions, we used VOSviewer version 1.6.7[3] [76]. For other analyses of keywords, we used python 3.8 and excel 2017.

## 3. Results

### 1.3 Overall statistics and distributions

A total of 1280 documents matched our search query. One of the papers was retracted, so we excluded it from the list. These documents have been cited 7121 times in overall. So, the average number of citations per document is 5.56. Table 1 shows the distribution of the document types.

The document types with the most average citations per paper are reviews, conference papers, and articles (which aggregately constitute 96.24% of data) on average received 49.44, 6.31 and 4.04 citations respectively. Approximately 78.58% of citations were to documents published within the last three years (2017 to 2019).

TABLE 1. Documents types

| Document type | Number | Average citations per paper |
|---|---|---|
| Conference Paper | 854 | 6.31 |
| Article | 369 | 4.04 |

[1] https://www.scopus.com

[2] https://www.scimagojr.com

[3] https://www.vosviewer.com

| | | |
|---|---|---|
| Conference Review | 38 | 2 |
| Review[4] | 8 | 49.44 |
| Book Chapter[5] | 8 | 0.25 |
| Note | 1 | 1 |
| Book | 1 | 0 |

Figure 1 shows the publication trend of documents in this field. The first paper found in our documents set dates back to 2004 by Scarselli, et al. [77] published in Lecture Notes in Computer Science (including subseries Lecture Notes in Artificial Intelligence and Lecture Notes in Bioinformatics). Scarselli, Tsoi and Gori, as the authors of the paper, are among top authors of the field. They proposed an architecture similar to recursive neural networks in which each unit stores the current node state, and, when activated, it calculates the next state using its neighbor's states.

A few numbers of documents were published before 2017 (41 documents). Thereafter, research in this field blossomed, with a significant rise in the number of documents. It is clear that this field of research is very young and is attracting more and more attention. The average annual growth of the documents from 2017 to 2019 is ~447%. It shows that this field is very young, attracting more attentions. Note that the citations curve in Figure 1 shows the number of citations to the papers published in a particular year. For instance, the plot shows that the papers published in 2017 are cited around 1800 times by the papers published thereafter. We observe two local peaks in the citations curve before 2017: one in 2005 and the other in 2009. These are mainly due to Gori, et al. [78] and Scarselli, et al. [79] papers. There is a rise in the number of citations in 2016 due to Li, et al. [35]. The year 2017 is an important milestone with a significant increase in the number of citations. Some of the most cited papers of the field appeared in this year including [80] and Bronstein, et al. [81]. The year 2018 is the most cited year with some notable papers including Yan, et al. [82], Schlichtkrull, et al. [83], Ying, et al. [16] and Zhang, et al. [84]. More than 50% of documents were published in 2019 with some remarkable documents including AlQuraishi [85] Wang, et al. [86] and Zhang, et al. [45][6].

[4] List of reviews are provided in Table 6.

[5] Titles of the books and book chapters are provided in APPENDIX A. List of books and book chapters about GNN

[6] Note that the data gathering was done about the half of 2020.

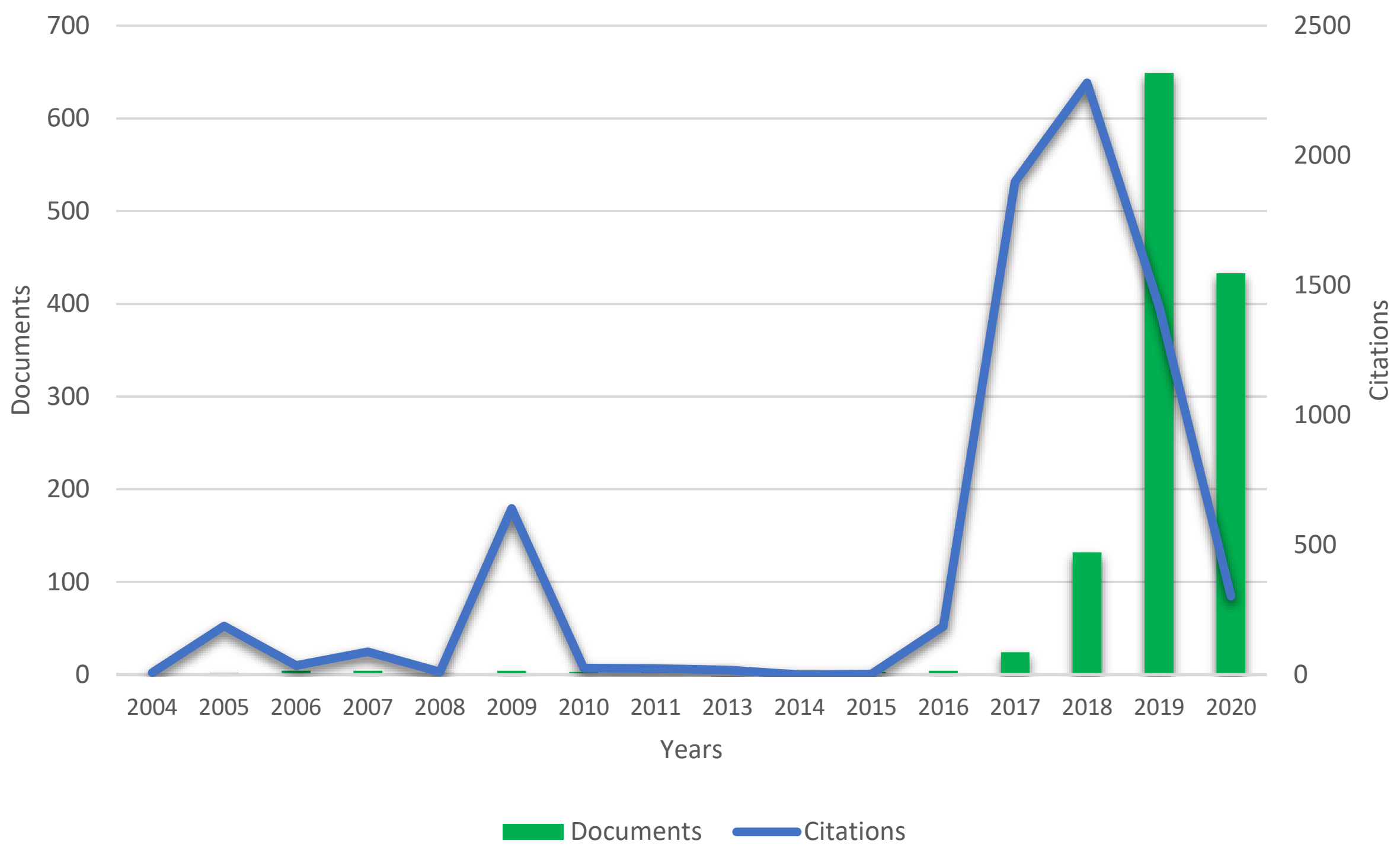

Figure 1. Yearly number of documents and citations

Figure 2 shows subject distribution of the GNN papers in the 22 different categories. The most frequent subjects are Computer Science (86%), Mathematics (22.90%), Engineering (22.59%), Decision Sciences (10.16%), Social Sciences (8.44%), Materials Science (5.16%), Physics and Astronomy (4.76%), Biochemistry, Genetics and Molecular Biology (3.98%), Business, Management and Accounting (3.67%), and Arts and Humanities (3.67%)[7].

[7] Note that a journal can be assigned to more than one category in Scopus.

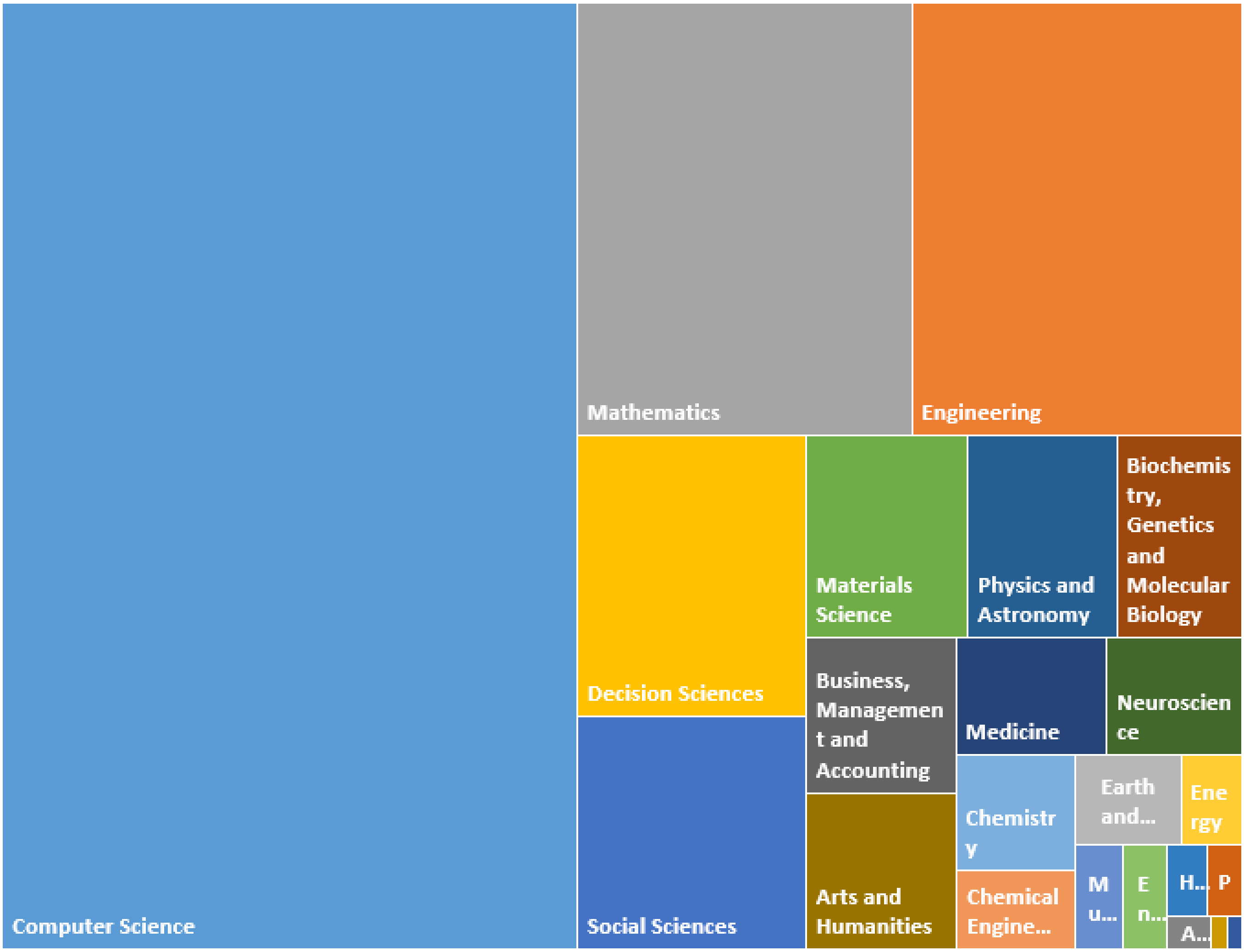

Figure 2. Different subject categories

### 2.3 Top authors

Table 2 shows the top ten authors based on their h-index [87] in this field[8]. Notice that the numbers reported in this table are limited to GNN papers which means the overall values can be larger. Franco Scarselli is the most prolific and impactful researcher in this field, with an h-index of 10. His top most frequently used keywords in descending order are ***graph neural networks***, ***graphical domains***, ***recursive neural networks,*** and ***deep neural networks***. He is an associate professor at Department of Information Engineering and Mathematics at the University of Siena. His main topics of research are ***Artificial intelligence, Machine learning, Artificial neural***

[8] Hirsch (2005) proposed the h-index. The H-index of a researcher is h if h of his/her papers have at least h citations each, and the other papers have at most h citations each.

***networks, Graph neural networks,*** and ***Deep learning***[9]. In addition, he has published two of the most-cited GNN papers which will be introduced in the Section F "Must-read papers".

Table 2. Most prolific and impactful researchers

| H-index | Author | All citations | Docs | Affiliation | Most used keywords |
|---|---|---|---|---|---|
| 10 | Franco Scarselli | 929 | 19 | University of Siena | Graph Neural Networks, Graphical Domains, Recursive Neural Networks, Deep Neural Networks |
| 8 | Marco Gori | 891 | 11 | University of Siena | Graphical Domains, Graph Neural Networks (GNNs), Biodegradability, Graph Processing |
| 8 | Wang Xiang | 252 | 33 | National University of Singapore | Graph Neural Network, Recommendation, Collaborative Filtering, Embedding Propagation |
| 8 | Markus Hagenbuchner | 732 | 14 | University of Wollongong | Graphical Domains, Recursive Neural Networks, Vapnik–Chervonenkis Dimension |
| 8 | Ah Chung Tsoi | 732 | 14 | University of Wollongong | Graphical Domains, Recursive Neural Networks, Approximation theory |
| 7 | Jian Tang | 165 | 21 | Syracuse University | Representation Learning, Network Embedding, Graph Neural Networks, Graph Convolutional Network, Graph Attention |
| 6 | Sanja Fidler | 204 | 7 | University of Toronto | Deep Learning, Grouping And Shape, Segmentation |
| 6 | Gabriele Monfardini | 830 | 6 | Università degli Studi di Siena | Graphical Domains, Graph Neural Networks (Gnns), Graph Processing, Relational Neural Networks |

[9] Based on his Google scholar profile.

| | | | | | |
|---|---|---|---|---|---|
| 5 | Jure Leskovec | 336 | 7 | Stanford University | Graph Neural Networks, Knowledge-Aware Recommendation, Label Propagation |
| 5 | Raquel Urtasun | 142 | 8 | University of Toronto | Graph Neural Networks, Inference, Message-Passing, Probabilistic Graphical Models |
| 5 | Ivan Titov | 356 | 5 | Universiteit van Amsterdam | – |
| 5 | Michael Bronstein | 767 | 8 | University of Lugano | Graph Convolutional Neural Networks, Geometric Deep Learning, Graph Neural Networks, Recommender Systems, Geometric Deep Learning |

### 3.3 Scientific collaboration

Studying the co-authorship patterns shows that each paper has 5.73 authors on average. Figure 3 shows the distribution of the number of authors and the average number of citations per paper. Sügis, et al. [88] authored the most collaborative paper with 18 authors. We can see that the number of citations increases from single-author papers to double-author papers, after which we observe a decrease in the average citations per paper except for the five-author papers. Collaborative papers (written by two or more authors) have been cited more than single-author papers on average (5.75 vs 3.82).

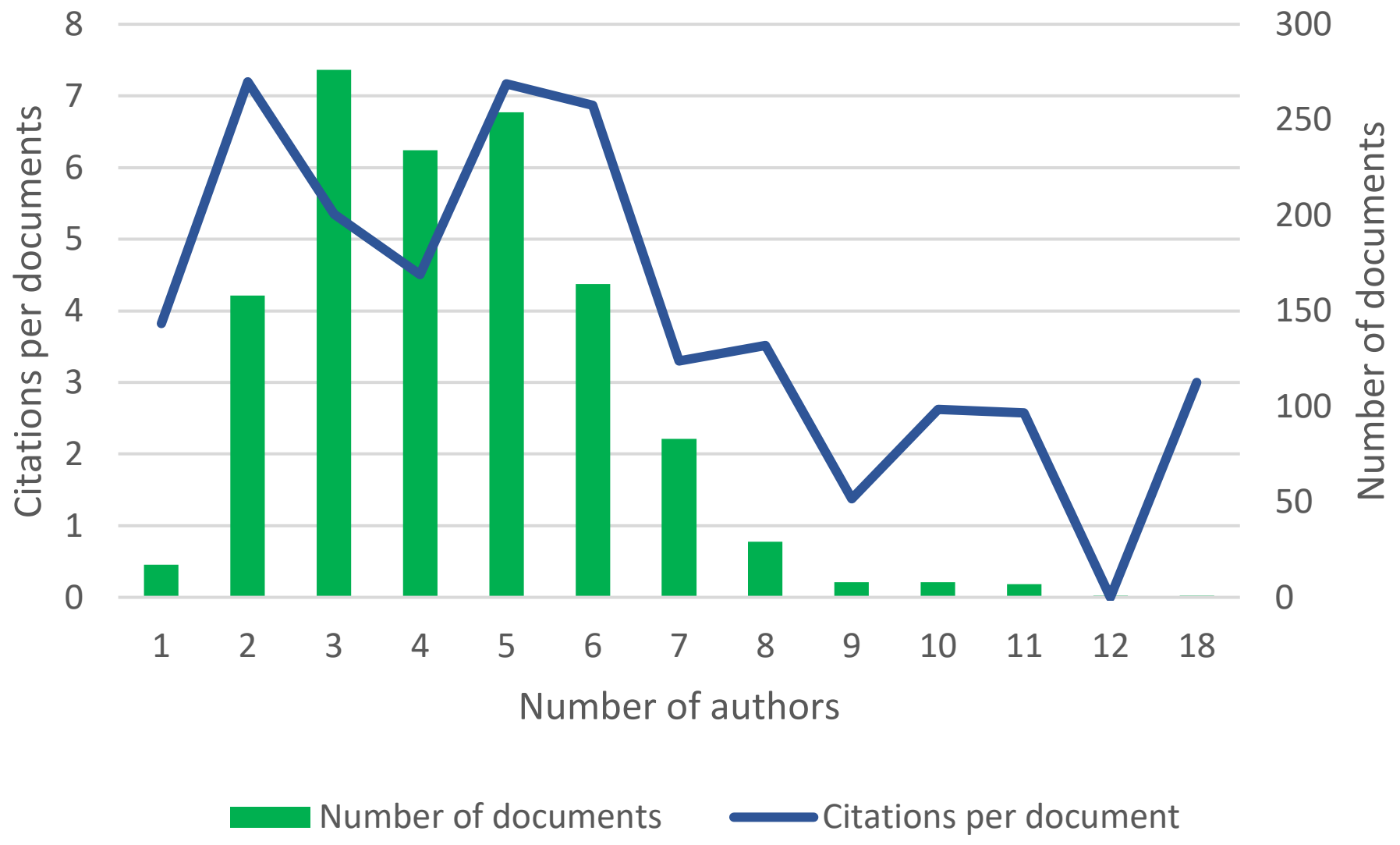

Figure 3. Distribution of the number of authors and the average number of citations per document

## 4.3 Top countries and institutions

A total of Fifty-one countries contributed to writing GNN documents. China (with 593 documents) published the greatest number of documents in this field, followed by the United States (with 377 documents), and Canada (with 82 documents). **Error! Reference source not found.** shows collaboration map of the most prolific countries[10]. The size of the nodes in the graph shows the number of documents published by the respective country. The graph edges show co-authorship and the node's colors indicate node clusters. Clustering has been done using VOS algorithm [89] based on collaborations. There are six clusters which can be explained partly by geographical distribution. Cluster one includes European countries, such as Netherlands, Belgium, and Spain. Cluster two has more diversity with three Asian countries plus the United States and Germany. Cluster three consists of East Asian, countries such as China, Japan, and South Korea. Cluster four consists of an East Asian country (Hong Kong), Australia and Italy which transfers knowledge between them and Europe. Also, it is clear that the United States acts as a bridge between China and European countries. Geographical patterns are not too clear in the Cluster five which consists of one Southeast Asian country (Singapore) one Middle East country (Israel) and one Northwestern Europe country (Switzerland). Cluster six consists of one Northwestern Europe country (Ireland) and Canada.

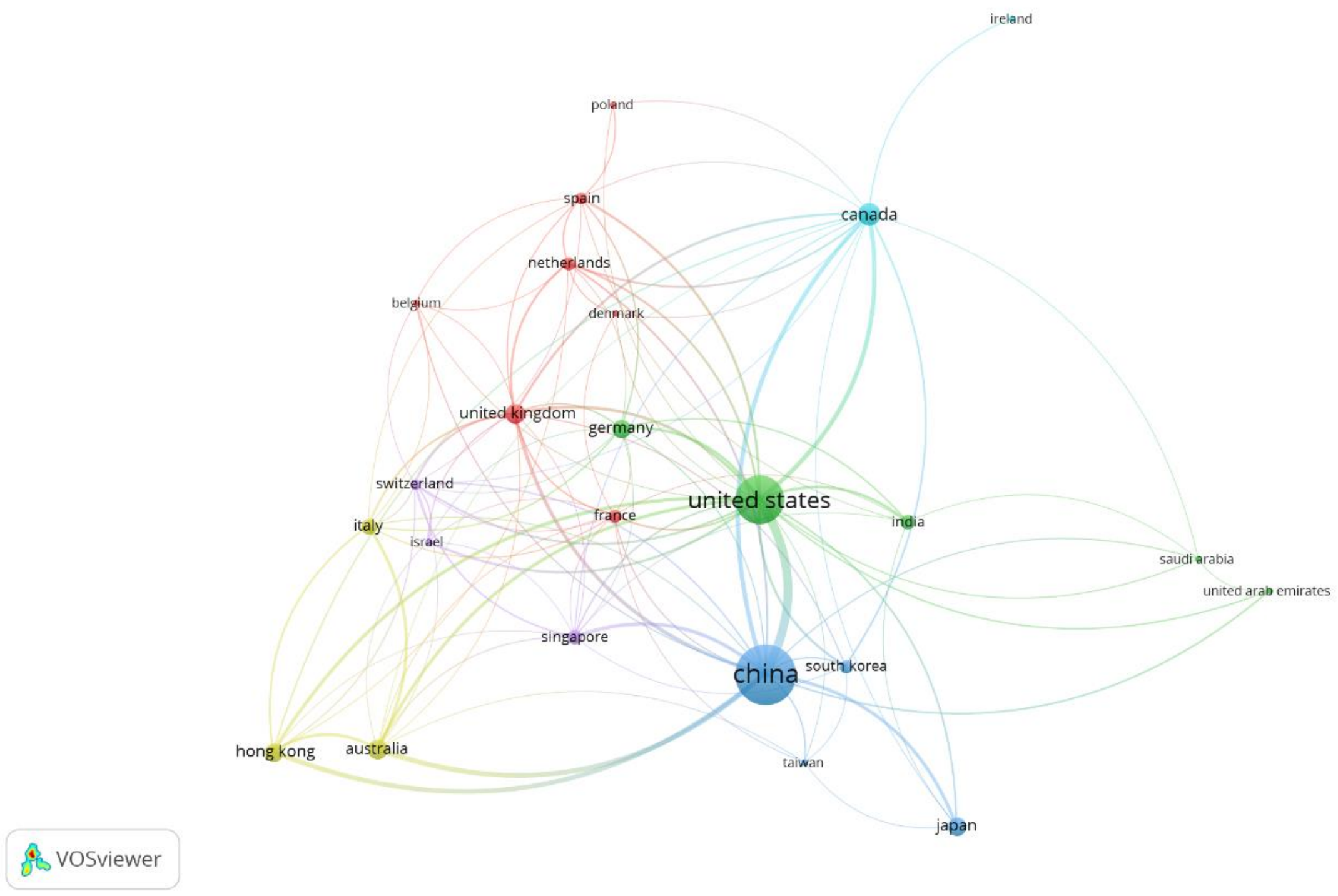

Figure 4. Top countries scientific collaboration

Table 3 shows the top ten prolific and impactful institutions in this field. Chinese institutions have published many documents, as expected because of China's population and its science productivity. There is just one Italian institution among the top ten most prolific institutions. On the other hand, the top-cited list is dominated by

[10] Twenty-four countries with more than five documents have been included in the map.

European and American institutions. University of Amsterdam is the most cited institution. There are some highly influential researchers at this university, such as Thomas Kipf, Max Welling, and Ivan Titov, who have contributed to important models and published highly cited papers. It is an interesting point that the Facebook research has appeared among the most influential institutions.

Table 3. Top institutions

| Most prolific | | | Most impactful | | |
|---|---|---|---|---|---|
| Institution | Country | Docs | Institution | Country | Citations |
| Chinese Academy of Sciences | China | 87 | University of Amsterdam | Netherlands | 975 |
| University of Chinese Academy of Sciences | China | 67 | University of Siena | Italy | 944 |
| Tsinghua University | China | 43 | Canadian Institute for Advanced Research | Canada | 767 |
| Peking University | China | 41 | University of Wollongong | Australia | 714 |
| Beijing University of Posts and Telecommunications | China | 31 | Hong Kong Baptist University | Hong Kong | 651 |
| Institute of Automation Chinese Academy of Sciences | China | 29 | New York University | United states | 492 |
| Beihang University | China | 29 | Universita della Svizzera Italiana | Italy | 459 |
| Tencent | China | 28 | Facebook research | United states | 435 |
| Shanghai Jiao Tong University | China | 23 | Swiss Federal Institute of Technology in Zurich | Switzerland | 435 |
| Università degli Studi di Siena | Italy | 23 | Université catholique de Louvain | Belgium | 435 |

### 5.3 Top publication sources

Table 4 shows the top ten publication sources which have published the greatest number of documents in this field. To evaluate these sources, we also included SJR and impact factor of these sources in the table. ***Lecture Notes in Computer Science Including Subseries Lecture Notes in Artificial Intelligence and Lecture Notes in Bioinformatics*** have published ~10.94% of the papers. ***Proceedings Of The IEEE International Conference On Computer Vision*** is the most impactful source in this table. Six of these sources are conferences, three are journals, and one is a book series. These 10 sources published around 27% of the documents in this field. In order to show the focus areas of these sources, we provided the most-used keywords in the fifth column. In the following, we provide a brief review of the most impactful papers of these journals based on the most frequent keywords.

Table 4. Top Sources

| Sources | Docs | SJR 2019 | IF 2019 | Most used keywords |
|---|---|---|---|---|
| Lecture Notes In Computer Science Including Subseries Lecture Notes In Artificial Intelligence And Lecture Notes In Bioinformatics | 140 | 0.42 | _ | Graph convolutional network, Representation learning, Knowledge graph |
| IEEE Access | 43 | 0.77 | 3.74 | Graph neural network, Deep learning, Link prediction |
| Proceedings Of The IEEE Computer Society Conference On Computer Vision And Pattern Recognition | 32 | 13.39 | 10.25 | Categorization, Deep learning, Knowledge graph |
| International Conference On Information And Knowledge Management Proceedings | 25 | 0.51 | _ | Recommender system, Heterogeneous graph, Link prediction |
| Proceedings Of The ACM SIGKDD International Conference On Knowledge Discovery And Data Mining | 25 | 1.004 | _ | LSTM, Graph embedding, Heterogeneous graph |
| Proceedings Of The IEEE International Conference On Computer Vision | 22 | 13.63 | _ | _ |
| Neurocomputing | 19 | 1.17 | 4.43 | Relational learning, Attention mechanism, LSTM |
| ACM International Conference Proceeding Series | 14 | 0.2 | _ | Auto-encoder, Skeleton-based, Action recognition |
| Knowledge Based Systems | 13 | 1.75 | 5.92 | Aspect-level, Recommender system, Graph neural network |
| Communications In Computer And Information Science | 12 | 0.188 | _ | Attention mechanism, Network embedding, Graph convolutional network |

**Graph convolutional network** (GCN)[80], as a special kind of GNNs, bridges the gap between spatial and spectral methods. The GCN propagation rule is equivalent to aggregating node representation with the direct neighbor node representations [90]. Clearly, node representations can be enriched by second-order neighbors' representations through adding another GCN layer. Yang, et al. [52] proposed Graph R-CNN for Scene Graph Generation. They used attentional GCN to integrate contextual information from neighboring objects in the scene. Also GCN has been used to model the spatial and semantic connections between objects for image captioning. Yao, et al. [48] leveraged this idea and proposed GCN-LSTM which uses LSTM with attention mechanism for sentence generation. It is interesting that GCN has opened its way into bioscience, too. Gievska and Madjarov [91] leveraged a modified version of GCN for prediction of the protein's functions based on their structure which is essentially modeled by a graph.

**Representation learning,** which mainly means graph representation learning in this context, aims to learn a low-dimensional vector representation of nodes (or edges) for graph data mining [32]. The learned vectors can be used in different downstream tasks such as node classification [75], [92], link prediction [92] , and community discovery [93]. Deepwalk [94] and node2vec [95] are examples of node representation learning. Sun, et al. [96] proposed a representation learning method which incorporates the entity descriptions embeddings (built by Doc2Vec [97]), with translation-based models for Medical Knowledge Graphs Representation Learning.

Another frequent keyword is **knowledge graph.** Formally, a knowledge graph is a collection of triplet (h, r, t) where h and t are head and tail entities, and r represents the relation from h to t [98]. The most popular knowledge graphs are WordNet, NELL, DBpedia, Freebase Google's Knowledge Graph and YAGO, which empowered different Natural Language Processing tasks including relation extraction, named entity recognition, and question answering [99]. The knowledge graphs have been used as a dataset for testing the models [100]. They have also been exploited to mine the relationship between classes for zero-shot recognition [101].

**Graph Neural Network** has been used in the context of Internet of Things (IoT), too. Zhang, et al. [102] represented the IoT system as a complete graph and proposed GNN-based Modeling approach for IoT (GNNM-IoT) which modeled the relationships between sensors with GNN. Yin, et al. [103] modeled interaction data in recommender systems by a bipartite user–item graph, and used some message passing layers to improve the latent factors of users and items.

**Deep learning** as a super field of GNN, has also been used by the authors as a keyword. Shi, et al. [51] proposed a GCN-based model for skeleton-based action recognition which consists of two types of graphs. One represents the common pattern for all the data, and the other represents the unique pattern of each type of data. The structure of these graphs are trained with convolutional parameters. Chen, et al. [53] proposed Multi-Label image recognition with Graph Convolutional Networks (ML-GCN) for multi-label image recognition. The idea was to model the label dependencies based on the objects co-occurrences in the images and to use a GCN to map the label graph into inter-dependent object classifiers.

**Link prediction** is the task of inferring upcoming likely interactions between nodes, given the existing graph [92]. GNNs has been popular tool for link prediction. Tan, et al. [104] proposed Combination-based knowledge Embedding model (CombinE), a knowledge graph embedding method which jointly minimized the norm of difference between entities' plus/minus combinations and the relation. Jing, et al. [32] introduced Variable Heat Kernel Representation (VHKRep) for graph representation learning which captures implicit global features by a heat diffusion kernel. They showed the effectiveness of their method on link prediction and node classification. Since the knowledge graphs are incomplete, a line of research focused on learning the knowledge graph representation (previously defined in this section).

**LSTM** (Long Short-Term Memory), which is a powerful sequence modeling method has been used as a rival for GNNs in flood prediction task [105]. Lu, et al. [106] combined GCN in LSTM cell and called it Graph LSTM (GLSTM) to handle graph sequences rather than sequential vectors for road speed prediction task.

**Graph embedding** and **Network embedding**, which have been used interchangeably, are among the most frequent keywords, too. Graph embedding is an effective method to represent graph data in a low dimensional space for graph analytics [107]. Hou, et al. [108] proposed a model named Property Graph Embedding (PGE) which incorporates nodes and edges properties into the procedure of graph embedding. Zhang, et al. [109], proposed Heterogeneous Graph Neural Network (HetGNN) based on the idea of leveraging heterogeneous structural and heterogeneous content information simultaneously.

**Relational learning,** refers to the learning paradigm where there may be relationships between examples or the examples may have an internal structure [110]. Interestingly, Trentin and Di Iorio [111] modeled the problem of graph classification in the form of Bayesian maximum-a-posteriori. Specifically, they calculated the class probability of the graph by multiplication of the class prior probability to the conditional probability of the graph relations.

**Attention mechanism** enables a model to focus on the most relevant parts of the input [112]. In the graph context, attention is defined as a function which assigns a relevance score [0, 1] to each of the nodes' neighbors. This score specifies the amount of attention the model gives to a particular neighbor [113]. Xie, et al. [114] proposed Attention-based Graph Convolution Networks (AGCN) for point clouds learning. They modeled the learning as a message propagation algorithm among adjacent points. Essentially, the model had three parts: local structural feature learning, point attention layer, and the global point network. They used the attention mechanism to model the relationship among k adjacent points.

Graph **auto-encoders** map graphs to low-dimensional vectors [115]. Liu and Sabbata [12] utilized Variational graph auto-encoders [116] to predict tweet geolocations. The model predicted the link between an unknown tweet and the existing tweet. Wang, et al. [115] proposed a training strategy to improve the training performance of graph auto-encoders. They injected noise to the adjacency matrix and used the noisy input to replace the input and the output.

**Skeleton-based** and **action recognition** are two keywords which co-occurred three times. In order to capture joint dependencies for action recognition, recent methods have constructed a skeleton graph whose vertices and edges are joints and bones, respectively. They applied GCN to extract correlated features [20]. More recently, Ding, et al. [19] proposed Semantics-guided Graph Convolutional Network (Sem-GCN). In order to aggregate information of the L-hop joint neighbors, the architecture utilized three semantic graph modules including structural graph extraction, actional graph inference and attention graph iteration. Yang, et al. [18] presented an end-to-end generative GCN to learn the joints graph connection from data. The model used self-attention to construct the weighted spatial graph of skeleton frames.

**Categorization** or classification as a fundamental task in machine learning refers to the process of predicting the class of a given sample. Node classification, as one of the basic graph analysis tasks, is usually performed to test the GNNs [117]. Li, et al. [20] proposed Actional-Structural Graph Convolution Network (AS-GCN) which stacked actional-structural and temporal graph convolution for action recognition. The structural links specified by the bones physical structure and the collaborative moving joints specified actional links. Kim, et al. [118] proposed Edge-Labeling Graph Neural Network (EGNN) which utilized a deep network for edge-labeling few-shot learning. For updating the nodes, the model aggregated features from inter/intra class neighbors of each node. After L updates, the edge label can be predicted based on the final edge feature.

A **recommender system** aims to provide personalized product or service recommendations for users in order to manage the growing information [119]. In this context, GCN has been used for Click-Through Rate (CTR) prediction [120], session-based recommendation [121], and agent-initiated recommendation [120].

**Heterogeneous graph** is another frequent keyword which refers to a kind of graph with more than one type of nodes or edges [122]. Li, et al. [123] constitute a heterogeneous graph composed of six kinds of nodes and eight kinds of edges for cross-domain aspect detection. Li, et al. [123] proposed GCN-based Anti-Spam (GAS) model composed of a heterogeneous graph to capture both the local and global contexts of a comment. Liu, et al. [124] introduced Graph Embeddings for Malicious accounts (GEM) i.e., for detecting malicious accounts which operated on account-device heterogeneous graph.

**Aspect-level** sentiment analysis, which is a subtask of sentiment analysis, aims to discover sentiments about entities, such as laptop and their aspects, such as battery life [36, 125]. Zhou, et al. [9] utilized Syntax- and Knowledge-based Graph Convolutional Network (SK-GCN). In order to enhance the sentence representation with respect to the given aspect, they leveraged syntactic dependency tree and commonsense knowledge graph using two GCNs. Zhao, et al. [11] utilized bidirectional attention mechanism with position encoding to model aspect-specific representations between each aspect and context words then exploited GCN over these representations to capture the sentiment dependencies between aspects in one sentence.

### 6.3 Must-read papers

Citation count is considered as an effective measure of the impact of a research paper [126-129]. In this section, we review the most-cited papers. We also present a list of available review papers and their suggested future directions and issues.

Table 5 shows ten papers with the greatest number of citations[11]. It should be noted that the third paper in this table is a review paper which is also included in Table 6. Since our purpose in this section is to review the main ideas of these hot papers, we ignore this review paper. Interestingly, eight out of ten papers are conference papers, which indicates the relative importance of conferences compared to journals in this field.

The most impactful paper is an article published in IEEE Transactions on Neural Networks, which changed its title in 2011. The current retitled publication is ***IEEE Transactions on Neural Networks and Learning Systems***, with an impact factor of 2.633, making it a prestigious journal in the field of deep learning. In this paper, Scarselli, et al. [79] proposed an architecture with forward and backward components. In the forward phase, the model computes states as a function of the target node's features, target node neighbors' features, the previous states, and the features of the edges which are connected to the target node. It stops when the difference between two states is less than or equal to a threshold. In the backward phase, the model computes gradient of a quadratic loss with respect to the model parameters. Indeed, they extended the framework of Gori, et al. [78] by conditioning the message-passing updates on initial edge features.

[11] The list of APPENDIX is provided in APPENDIX D. APPENDIX

The second paper is the famous GCN paper of Kipf and Welling [80], a revolutionary paper in this field that combines the two different approaches. Basically, the model combines each node representation with its direct neighbors in each layer.

Monti, et al [26] proposed Mixture Model Networks (MoNet), a special model to extend CNNs to graphs and manifolds. The model associates each neighbor of the point (x) with a d-dimensional pseudo-coordinates vector u (x, y). It then, applies a set of Gaussian kernels with some learnable parameters to these coordinates, instead of using fixed kernels. Yan, et al. [82] presented Spatial-Temporal Graph Convolutional Networks (ST-GCN) for action recognition which learns both the spatial and temporal patterns using GNN. The spatio-temporal graph is constructed by both intra-body edges of joints based on the natural connections and inter-frame edges which connect the same joints in the neighboring frames. Li, et al. [35] introduced Gated Graph Neural Networks (GG-NNs). This model, as a modification of graph neural network (first paper in Table 5 [79]) uses gated recurrent units (GRUs) to generate sequences. Gori, et al. [78] proposed a neural network model which directly acted on graphs. The system computes the state of the node ***n*** ($x_n$) as a function of its features along with the features and states of its neighbors. Schlichtkrull, et al. [83] proposed Relational Graph Convolutional Networks (R-GCNs) for multi-graphs and evaluated the model on link prediction and entity classification tasks. The architecture of the model is quite simple in each layer, the representation of each node is obtained by a combination of that node and its neighbors in different graphs. Ying, et al. [16] developed PinSage to generates the node embeddings for web-scale recommendation. The architecture computes the target node's embedding based on its pervious representation and the representation of its neighbors which are computed based on their neighbors. In contrast to the mainstream of GCNs which are based on the powers of the graph Laplacian, PinSage performs by sampling the target node's neighborhood. Finally, Marcheggiani and Titov [42] utilized GCN over syntactic dependency trees as sentence encoder for semantic role labeling. Their experiments showed that stacking GCN and LSTM layers outperformed the state-of-the-art on CoNLL-2009.

Table 5. Top ten cited papers

| Title | Citation | Authors | Year | Document type | Source |
|---|---|---|---|---|---|
| The graph neural network model | 578 | Scarselli, F.; Gori, M.; Tsoi, A.C.; Hagenbuchner, M.; Monfardini, G. | 2009 | Article | IEEE Transactions on Neural Networks |
| Semi-supervised classification with graph convolutional networks | 577 | Kipf, T.N.; Welling, M. | 2017 | Conference Paper | 5th International Conference on Learning Representations, ICLR 2017 |
| Geometric Deep Learning: Going beyond Euclidean data | 435 | Bronstein, M.M.; Bruna, J.; Lecun, Y.; Szlam, A.; Vandergheynst, P. | 2017 | Review | IEEE Signal Processing Magazine |
| Geometric deep learning on graphs and manifolds using mixture model CNNs | 214 | Monti, F.; Boscaini, D.; Masci, J.; RodolÃ , E.; Svoboda, J.; Bronstein, M.M. | 2017 | Conference Paper | 30th IEEE Conference on Computer Vision and Pattern Recognition, CVPR 2017 |
| Spatial temporal graph convolutional networks for skeleton-based action recognition | 166 | Yan, S.; Xiong, Y.; Lin, D. | 2018 | Conference Paper | 32nd AAAI Conference on Artificial Intelligence, AAAI 2018 |
| Gated graph sequence neural networks | 161 | Li, Y.; Zemel, R.; Brockschmidt, M.; Tarlow, D. | 2016 | Conference Paper | 4th International Conference on Learning Representations, ICLR 2016 |
| A new model for learning in graph domains | 159 | Gori, M.; Monfardini, G.; Scarselli, F. | 2005 | Conference Paper | International Joint Conference on Neural Networks, IJCNN 2005 |
| Modeling Relational Data with Graph Convolutional Networks | 147 | Schlichtkrull, M.; Kipf, T.N.; Bloem, P.; van den Berg, R.; Titov, I.; Welling, M. | 2018 | Conference Paper | 15th International Conference on Extended Semantic Web Conference, ESWC 2018 |
| Graph convolutional neural networks for web-scale recommender systems | 136 | Ying, R.; He, R.; Chen, K.; Eksombatchai, P.; Hamilton, W.L.; Leskovec, J. | 2018 | Conference Paper | 24th ACM SIGKDD International Conference on Knowledge Discovery and Data Mining, KDD 2018 |
| Encoding sentences with graph convolutional networks for semantic role labeling | 107 | Marcheggiani, D.; Titov, I. | 2017 | Conference Paper | 2017 Conference on Empirical Methods in Natural Language Processing, EMNLP 2017 |

Review papers are good starting points for those who want to work on a field and identify research gaps. Table 6 shows the list of reviews conducted in the field of GNN. Bronstein, et al. [81] is the most cited review paper. The other seven review papers are all published in 2019 or 2020.

Table 6. GNN review papers

| Title | Year | Citations | Suggested future directions and open problems |
|---|---|---|---|
| Geometric Deep Learning Going beyond Euclidean data 77 | 2017 | 435 | 1. Generalization across different domains. |
| | | | 2. Dealing with signals over dynamic structures. |
| | | | 3. Coping with directed graphs. |
| | | | 4. Learning generative models. |
| | | | 5. Developing efficient computational paradigms. |
| An Overview of Unsupervised Deep Feature Representation for Text Categorization 126 | 2019 | 5 | 1. Exploring more efficient unsupervised deep learning models. |
| A gentle introduction to deep learning for graphs 127 | 2020 | 3 | 1. Formalizing various adaptive graph processing techniques under a unified framework. |
| | | | 2. Defining a set of benchmarks in order to assess proposed models. |
| | | | 3. Transferring research knowledge to other application fields. |
| Graph convolutional networks for computational drug development and discovery 128 | 2020 | 2 | 1. Extending gcns to 3D structures (such as Molecular compounds). |
| | | | 2. Exploring motif-based GCN and its application on drug discovery. |
| | | | 3. Defining convolution on hyper graphs (for example drugs with the same adrs, targets or indications). |
| Introduction to Graph Neural Networks 129 | 2020 | 0 | 1. Deepening GNNs regarding over smoothing problem. |
| | | | 2. Dealing with dynamic networks. |
| | | | 3. Generating optimal graphs for non-structured data. |
| | | | 4. Applying embedding models in the web-scale. |
| Learning Combinatorial Optimization on Graphs: A Survey with Applications to Networking 130 | 2020 | 0 | 1. Improving scalability, adaptability, generalization, and run time of gnns. |
| | | | 2. Automating above improvements without re-training. |
| | | | 3. Using distributed machine learning. |
| Application of deep learning in ecological resource research: Theories, methods, and challenges 131 | 2020 | 0 | 1. Standardizing and sharing of data for ecological resource research. |
| | | | 2. Increasing ability to explain hidden layers |
| | | | 3. Appling more advanced deep learning methods on ecological resource research. |

## 7.3 Keyword analysis

### 7.3.1 The Most used keywords

We analyze the most frequently used keywords in two time spans, 2004-2017 and 2017-2020. The word clouds of these periods are illustrated in Figure 5. Word clouds are used to visually summarize texts [130]. The size of keywords in the word clouds indicates their frequency in the respective time period. ***Graph neural network, relational learning, structured pattern recognition, recurrent network, feedforward network, neuroscience, random process, recursive neural network, graph structured data, semigraph, wavelet transform,*** *and* ***graphical domain*** are the most used keywords in the first period. In the second period ***graph convolutional network, graph neural network, deep learning, geometric deep learning, representation learning, machine learning, graph convolution, network embedding, convolutional neural network, knowledge graph, neural network, and action recognition*** are used most frequently by authors[12]. As it is obvious from this figure, ***graph neural network*** and ***graph convolutional network*** are the most frequent keywords in the first and second periods, respectively. This is because, as mentioned previously, early GNNs were more similar to RNNs with different states in different steps, while the recent models are more convolutional-based. Also, some topics, such as recursive neural network, have lost their positions over time, which demonstrates the change of approach to deep learning on graphs. Some technical topics have emerged or grown in the recent years, including representation learning, graph attention, graph autoencoder, variational autoencoder, spectral graph theory, message-passing, graph isomorphism test, label propagation, and balance theory. While early GNNs where based on message-passing too, the message-passing used in more recent methods refers to recent advancements such as [80, 131]. GNNs have been applied for different applications including action recognition, semantic segmentation, anomaly detection, drug discovery, sentiment analysis, session-based recommendation, video analytics, scene graph generation, social recommendation, image captioning, human pose estimation, traffic forecasting, visual question answering, traffic speed prediction, name disambiguation, hyperspectral image classification, and knowledge graph completion.

[12] http://rank.sid.ir/cloud

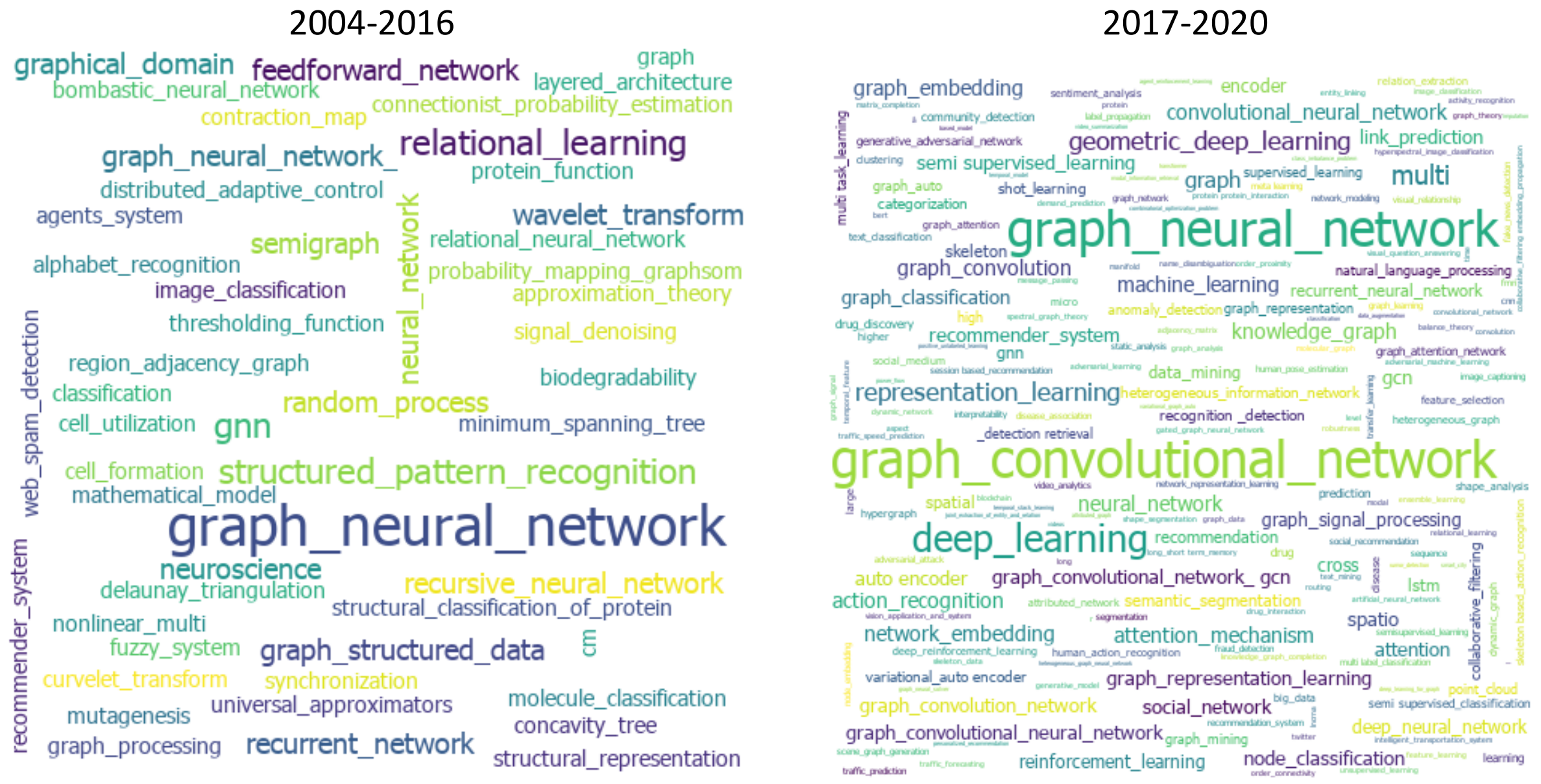

Figure 5. Word clouds of the keywords used in the of GNN research papers in the two different time span[13]

### 7.3.2 Hot topics

**Table 7** shows ten keywords with the highest average publication year to reveal topics that have received the most attention recently. In order to remove the possible noise, we include the keywords with more than two papers[14].

Table 7. Ten keywords with the highest average publication year (frequency > 2)

| keyword | Number of papers | Average publication year | Type |
|---|---|---|---|
| BERT | 3 | 2020 | Model |
| Dynamic network | 3 | 2020 | Network |
| Graph attention network | 6 | 2020 | Model |

[13] List of keywords is provided in APPENDIX .

[14] For the continue of keywords with the highest average publication year refer to APPENDIX 3.

| | | | |
|---|---|---|---|
| Relation extraction | 5 | 2019.8 | Task |
| Attention mechanism | 21 | 2019.8 | Model |
| Human pose estimation | 4 | 2019.8 | Task |
| Self-supervised learning | 4 | 2019.8 | Learning approach |
| Semisupervised learning | 4 | 2019.8 | Learning approach |
| Traffic prediction | 4 | 2019.8 | Task |
| Adversarial learning | 3 | 2019.6 | Learning approach |

The first keyword is **BERT,** a successful language-understanding model based on transformer [132] which has been used recently in this field for token representation. Jeong, et al. [43], used BERT as the encoder of context sentences and GCN as citation context encoder in the task of context-aware paper recommendation.

**Dynamic network** refers to a sequence of graph snapshots over time. Mahdavi, et al. [133] proposed Dynamic joint Variational Graph Auto-Encoders (Dyn-VGAE) which consists of auto-encoders that embed graph snapshots based on their local structures and interact with each other to learn temporal dependencies of graphs.

**Graph attention network** specifies different weights for aggregating different neighbors [112] to obtain a weighted average of neighbors' features. In this way, the model can overcome the cross-class links. Zhao, et al. [98] used different aggregation functions for the task of Out Of Knowledge Graph (OOKG) entity and relation. They leveraged average pooling, max pooling, and attention as the aggregation functions.

**Relation extraction** is an NLP task which aims to distinguish relational facts from a piece of text. Xie, et al. [134] proposed a GNN with a propagation rule similar to GCN [80] on the heterogeneous graph composed of sentence and entity nodes for few-shot relation classification.

**Attention mechanism,** introduced previously, is also among new topics. You, et al. [135] proposed Sliced recurrent neural network and Attention treated GCN-based Parallel (SAGP) model for remote sensing image recognition which is composed of two sub-modules: the improved Sliced Recurrent Neural Network (SRNN) retains the semantic information of the context and the original image features and a GCN which mines high-weight features (obtained by attention mechanism) and reserves the relationship between their features.

**Human pose estimation's** goal is to identify the human body parts poses in images or videos [136]. Wang, et al. [137] proposed to utilize Global Relation Reasoning Graph Convolutional Networks (GRR-GCN) to model the global dependencies of body joints. The model projects the coordinate space features to a fully-connected graph, in which global relation reasoning is done by GCN.

Bin, et al. [138] proposed a model which first feeds images to CNNs in order to obtain the key points representations. The model has two parallel multi-layer Pose Graph Convolutional Network (PGCN) modules which then capture the feature correlation between key points locally and non-locally based on a directed graph over the obtained key representations.

**Self-supervised learning** is an emerging effective learning strategy which creates a supervised task from unlabeled data. For instance, the model can learn to predict half of an image given the other half [139, 140]. Shen, et al. [141] used self-supervised learning for taxonomy expansion data generation. They suggested TaxoExpan a GCN-based neural network to learn to predict if a query concept is the hyponym of an anchor concept. Bo, et al. [142] proposed Structural Deep Clustering Network (SDCN) which uses a delivery operator in order to combine representations of auto-encoders and GCN layers. In effect, it leverages a dual self-supervised strategy to unify these deep learning models. **Semisupervised learning** is an approach to machine learning in which only a small subset of training samples are labeled [80]. The goal is to infer the labels of unlabeled samples from the information contained in feature vectors and labeled samples [143]. Qin, et al. [50] proposed Spectral–Spatial Graph Convolutional Networks (S2GCNs) for Hyperspectral Image Classification (HIC) which is a semisupervised GCN based model that utilizes spatial (pixel adjacency) and spectral information.

**Traffic prediction** is the task of forecasting real-time traffic based on floating car and historical data including flow, average speed, and incidents [144]. Zhao, et al. [145] proposed SpatioTemporal Data Fusion (STDF) for traffic prediction which separates data into traffic directly/indirectly-related data. The model leverages GCN for processing directly-related data.

**Adversarial learning,** the final hot topic, is a learning technique that tries to fool algorithms by presenting deceptive inputs to them. Hong, et al. [146] proposed an architecture based on Generative Adversarial Network (GAN) to predict missing longitudinal diffusion MRI data. They leveraged graph convolution in both generator and discriminator of the network.

## 4. Conclusion

In this paper, we presented a bibliometric overview of the young and fast-growing field of GNNs. The publication trend in the field shows that GNNs has recently attracted a huge amount of attention from scientific community. Moreover, GCN has recently become a more popular term than GNN. Conference papers constitute the main source of impactful publications in this field. Franco Scarselli, one of the authors of the first GNN paper, is the most active and impactful researcher in the field with the highest h-index.

The most active institutions are from China and the most impactful institutions are mostly from Europe and the United States. Lecture Notes in Computer Science including subseries Lecture Notes in Artificial Intelligence and Lecture Notes in Bioinformatics has published the most number of GNN papers.

Based on the most used keywords, the node classification is the most popular task, followed by link prediction, and graph classification. In addition, it can be inferred from the recent keywords that the *attention mechanism* is now an important topic. The reason is that the GCNs mainly averages the representations of all the neighbors without considering their features' similarity, which in turn can bring noise to node representations.

It should be noted that the presented results are limited in some senses. First, we ignored the papers which are not indexed by Scopus. Second, even though we covered the main phrases in the field for data collection, it is possible that the list is not comprehensive. Third, some fields of records in Scopus were empty, which could affect the results. As a future work, we can provide a taxonomy of tasks which have been tackled using GNNs and their results.

## Appendices

### APPENDIX A. List of books and book chapters about GNN

| Title | Type | Year | Source |
|---|---|---|---|
| Riemannian Geometric Statistics in Medical Image Analysis [147] | Book | 2019 | Riemannian Geometric Statistics in Medical Image Analysis |
| Efficient Data Augmentation Using Graph Imputation Neural Networks [148] | Book Chapter | 2021 | Smart Innovation, Systems and Technologies |
| Unsupervised Learning Towards the Future [149] | Book Chapter | 2020 | Advances in Computer Vision and Pattern Recognition |
| A Neural Network Model to Include Textual Dependency Tree Structure in Gender Classification of Russian Text Author [46] | Book Chapter | 2020 | Mechanisms and Machine Science |
| Unsupervised Visual Learning: From Pixels to Seeing [150] | Book Chapter | 2020 | Advances in Computer Vision and Pattern Recognition |
| Coupling Appearance and Motion: Unsupervised Clustering for Object Segmentation Through Space and Time [151] | Book Chapter | 2020 | Advances in Computer Vision and Pattern Recognition |
| Graph Convolutional Networks on Customer/Supplier Graph Data to Improve Default Prediction [152] | Book Chapter | 2019 | Springer Proceedings in Complexity |
| Deep Neural Networks for Structured Data [153] | Book Chapter | 2018 | Studies in Computational Intelligence |
| Curvelet Interaction with Artificial Neural Networks [153] | Book Chapter | 2016 | Studies in Computational Intelligence |

### APPENDIX B. List of most frequent keywords (frequency > 3)

| Keywords | Frequency | Keywords | Frequency | Keywords | Frequency |
|---|---|---|---|---|---|
| graph neural network | 250 | big data | 5 | deep reinforcement learning | 5 |
| graph convolutional network | 247 | heterogeneous information network | 8 | attributed network | 5 |
| deep learning | 118 | categorization | 8 | clustering | 5 |
| geometric deep learning | 42 | attention | 8 | network representation learning | 4 |
| representation learning | 40 | recommendation | 7 | multi-label classification | 4 |
| machine learning | 28 | point cloud | 7 | network modeling | 4 |
| graph convolution | 28 | natural language processing | 7 | image captioning | 4 |
| neural network | 27 | multi-task learning | 7 | human pose estimation | 4 |
| network embedding | 27 | graph representation | 7 | hypergraph | 4 |
| convolutional neural network | 27 | graph mining | 7 | meta-learning | 4 |
| knowledge graph | 26 | anomaly detection | 7 | image classification | 4 |
| action recognition | 24 | variational autoencoder | 6 | interpretability | 4 |
| graph embedding | 23 | semi-supervised classification | 6 | label propagation | 4 |
| node classification | 21 | relational learning | 6 | fraud detection | 4 |
| attention mechanism | 21 | lstm | 6 | graph network | 4 |
| graph convolutional neural network | 20 | gnn | 6 | graph autoencoder | 4 |
| graph representation learning | 19 | generative adversarial network | 6 | fake news detection | 4 |
| graph | 18 | graph attention network | 6 | graph convolutional network (gcns) | 4 |
| semi-supervised learning | 17 | drug discovery | 6 | graph learning | 4 |
| social network | 17 | community detection | 6 | embedding propagation | 4 |
| recommender system | 16 | text classification | 5 | adversarial attack | 4 |
| link prediction | 16 | sentiment analysis | 5 | traffic prediction | 4 |
| graph convolution network | 15 | skeleton | 5 | structured pattern recognition | 4 |
| graph convolutional network (gcn) | 15 | shape analysis | 5 | transfer learning | 4 |
| graph classification | 13 | relation extraction | 5 | static analysis | 4 |
| reinforcement learning | 12 | skeleton-based action recognition | 5 | video analytics | 4 |
| deep neural network | 12 | social medium | 5 | visual relationship | 4 |
| gcn | 11 | prediction | 5 | session-based recommendation | 4 |
| graph signal processing | 11 | human action recognition | 5 | semisupervised learning | 4 |
| collaborative filtering | 11 | graph attention | 5 | skeleton-based | 4 |
| retrieval | 9 | graph embeddings | 5 | recommendation system | 4 |
| semantic segmentation | 9 | feature selection | 5 | social recommendation | 4 |
| recurrent neural network | 9 | few-shot learning | 5 | scene graph generation | 4 |
| data mining | 9 | heterogeneous graph | 5 | segmentation | 4 |
| autoencoder | 9 | dynamic graph | 5 | self-supervised learning | 4 |
| recognition: detection | 8 | | | | |

**APPENDIX C. List of keywords with the highest average publication year (frequency > 2). For the top ten keywords refer to Table 7).**

| keyword | Number of papers | Average publication year | Average citations | Average normalize citations | keyword | Number of papers | Average publication year | Average citations | Average normalize citations |
|---|---|---|---|---|---|---|---|---|---|
| artificial neural network | 3 | 2019.67 | 0 | 0 | point cloud | 7 | 2019.43 | 0.43 | 0.33 |
| blockchain | 3 | 2019.67 | 0 | 0 | recommendation | 7 | 2019.43 | 10.57 | 4.98 |
| convolutional network | 3 | 2019.67 | 2.33 | 1.07 | attributed network | 5 | 2019.4 | 4.4 | 1.21 |
| graph network | 3 | 2019.67 | 0.33 | 0.48 | deep reinforcement learning | 5 | 2019.4 | 1.6 | 1.9 |
| lstm | 6 | 2019.67 | 1 | 1.43 | feature selection | 5 | 2019.4 | 0 | 0 |
| protein-protein interaction | 3 | 2019.67 | 0 | 0 | graph attention | 5 | 2019.4 | 6 | 0.35 |
| robustness | 3 | 2019.67 | 0.67 | 0.31 | graph convolution network | 15 | 2019.4 | 1.47 | 0.8 |
| text mining | 3 | 2019.67 | 0 | 0 | skeleton-based action recognition | 5 | 2019.4 | 3.2 | 0.78 |
| traffic speed prediction | 3 | 2019.67 | 0.67 | 0.95 | text classification | 5 | 2019.4 | 0.8 | 0.37 |
| twitter | 3 | 2019.67 | 1 | 1.11 | graph classification | 13 | 2019.38 | 0.77 | 0.38 |
| visual question answering | 3 | 2019.67 | 0 | 0 | attention | 8 | 2019.38 | 0.5 | 0.35 |
| few-shot learning | 5 | 2019.6 | 0.4 | 0.18 | heterogeneous information network | 8 | 2019.38 | 0.25 | 0.11 |
| sentiment analysis | 5 | 2019.6 | 0.2 | 0.29 | demand prediction | 3 | 2019.33 | 0.67 | 0.95 |
| social medium | 5 | 2019.6 | 0.2 | 0.29 | dynamic graph | 3 | 2019.33 | 1.33 | 1.26 |
| gcn | 11 | 2019.55 | 0.09 | 0.04 | gated graph neural network | 3 | 2019.33 | 0 | 0 |
| fake news detection | 4 | 2019.5 | 0 | 0 | graph analysis | 3 | 2019.33 | 9.67 | 4.42 |
| graph learning | 4 | 2019.5 | 0.5 | 0.47 | graph data | 3 | 2019.33 | 0.33 | 0.15 |
| meta-learning | 4 | 2019.5 | 0.25 | 0.11 | graph signal | 3 | 2019.33 | 1.33 | 0.61 |
| session-based recommendatio n | 4 | 2019.5 | 5.5 | 2.76 | hyperspectral image classification | 3 | 2019.33 | 0.67 | 0.95 |
| skeleton-based | 4 | 2019.5 | 0 | 0 | knowledge graph completion | 3 | 2019.33 | 0 | 0 |
| node classification | 21 | 2019.48 | 0.19 | 0.09 | name disambiguation | 3 | 2019.33 | 0 | 0 |
| graph representation learning | 19 | 2019.47 | 1.11 | 0.66 | routing | 3 | 2019.33 | 3 | 0.17 |
| collaborative filtering | 11 | 2019.45 | 7.27 | 3.33 | semantic segmentation | 9 | 2019.33 | 1 | 0.8 |
| reinforcement learning | 11 | 2019.45 | 0.55 | 0.34 | Siamese network | 3 | 2019.33 | 0 | 0 |
| anomaly detection | 7 | 2019.43 | 2.14 | 0.98 | skeleton data | 3 | 2019.333 | 1 | 0.466 |

## APPENDIX D. Eleven to sixty most cited papers (for the top ten papers refer to Table 5)

| Number | Title | Citations | Authors | Year | Document type |
|---|---|---|---|---|---|
| 11 | Large-Scale Point Cloud Semantic Segmentation with Superpoint Graphs | 101 | Landrieu, L.; Simonovsky, M. | 2018 | Conference Paper |
| 12 | FastGCN: Fast learning with graph convolu-tional networks via importance sampling | 92 | Chen, J.; Ma, T.; Xiao, C. | 2018 | Conference Paper |
| 13 | An end-to-end deep learning architecture for graph classification | 86 | Zhang, M.; Cui, Z.; Neumann, M.; Chen, Y. | 2018 | Conference Paper |
| 14 | Modeling polypharmacy side effects with graph convolutional networks | 80 | Zitnik, M.; Agrawal, M.; Leskovec, J. | 2018 | Conference Paper |
| 15 | 3D Graph Neural Networks for RGBD Semantic Segmentation | 79 | Qi, X.; Liao, R.; Jia, J.; Fidler, S.; Urtasun, R. | 2017 | Conference Paper |
| 16 | Zero-Shot Recognition via Semantic Embeddings and Knowledge Graphs | 70 | Wang, X.; Ye, Y.; Gupta, A. | 2018 | Conference Paper |
| 17 | Deeper insights into graph convolutional networks for semi-supervised learning | 70 | Li, Q.; Han, Z.; Wu, X.-M. | 2018 | Conference Paper |
| 18 | Geometric matrix completion with recurrent multi-graph neural networks | 70 | Monti, F.; Bronstein, M.M.; Bresson, X. | 2017 | Conference Paper |
| 19 | Graph convolutional encoders for syntax-aware neural machine translation | 68 | Bastings, J.; Titov, I.; Aziz, W.; Marcheggiani, D.; Simaâ€™an, K. | 2017 | Conference Paper |
| 20 | Spatio-temporal graph convolutional networks: A deep learning framework for traffic forecasting | 67 | Yu, B.; Yin, H.; Zhu, Z. | 2018 | Conference Paper |
| 21 | Designing a new mathematical model for cellular manufacturing system based on cell utilization | 66 | Mahdavi, I.; Javadi, B.; Fallah-Alipour, K.; Slomp, J. | 2007 | Article |
| 22 | Graph attention networks | 63 | VeliÄkoviÄ‡, P.; Casanova, A.; LiÃ², P.; | 2018 | Conference Paper |
| | | | Cucurull, G.; Romero, A.; Bengio, Y. | | |
| 23 | Convolutional neural networks on surfaces via seamless toric covers | 57 | Maron, H.; Galun, M.; Aigerman, N.; Trope, M.; Dym, N.; Yumer, E.; Kim, V.G.; Lipman, Y. | 2017 | Conference Paper |
| 24 | Efficient Interactive Annotation of Segmentation Datasets with Polygon-RNN++ | 56 | Acuna, D.; Ling, H.; Kar, A.; Fidler, S. | 2018 | Conference Paper |
| 25 | Computational capabilities of graph neural networks | 51 | Scarselli, F.; Gori, M.; Tsoi, A.C.; Hagenbuchner, M.; Monfardini, G. | 2009 | Article |
| 26 | Protein interface prediction using graph convolutional networks | 50 | Fout, A.; Byrd, J.; Shariat, B.; Ben-Hur, A. | 2017 | Conference Paper |
| 27 | Large-scale learnable graph convolutional networks | 47 | Gao, H.; Wang, Z.; Ji, S. | 2018 | Conference Paper |
| 28 | End-to-End Differentiable Learning of Protein Structure | 45 | AlQuraishi, M. | 2019 | Article |
| 29 | Adversarial attacks on neural networks for graph data | 45 | ZÃ¼gner, D.; Akbarnejad, A.; GÃ¼nnemann, S. | 2018 | Conference Paper |
| 30 | Few-shot learning with graph neural networks | 44 | Garcia, V.; Bruna, J. | 2018 | Conference Paper |
| 31 | Link prediction based on graph neural networks | 44 | Zhang, M.; Chen, Y. | 2018 | Conference Paper |
| 32 | SplineCNN: Fast Geometric Deep Learning with Continuous B-Spline Kernels | 42 | Fey, M.; Lenssen, J.E.; Weichert, F.; Muller, H. | 2018 | Conference Paper |
| 33 | Neural graph collaborative filtering | 41 | Wang, X.; He, X.; Wang, M.; Feng, F.; Chua, T.-S. | 2019 | Conference Paper |
| 34 | Graph convolution over pruned dependency trees improves relation extraction | 40 | Zhang, Y.; Qi, P.; Manning, C.D. | 2020 | Conference Paper |
| 35 | Hierarchical graph representation learning with differentiable pooling | 39 | Ying, R.; Morris, C.; Hamilton, W.L.; You, J.; Ren, X.; Leskovec, J. | 2018 | Conference Paper |
| 36 | Graph-based neural multi-document summarization | 39 | Yasunaga, M.; Zhang, R.; Meelu, K.; Pareek, A.; Srinivasan, K.; Radev, D. | 2017 | Conference Paper |
| 37 | How powerful are graph neural networks? | 38 | Xu, K.; Jegelka, S.; Hu, W.; Leskovec, J. | 2019 | Conference Paper |
| 38 | Topology-Aware Prediction of Virtual Network Function Resource Requirements | 38 | Mijumbi, R.; Hasija, S.; Davy, S.; Davy, A.; Jennings, B.; Boutaba, R. | 2017 | Article |
| 39 | Adversarially regularized graph autoencoder for graph embedding | 37 | Pan, S.; Hu, R.; Long, G.; Jiang, J.; Yao, L.; Zhang, C. | 2018 | Conference Paper |
| 40 | Graph convolutional policy network for goal-directed molecular graph generation | 36 | You, J.; Liu, B.; Ying, R.; Pande, V.; Leskovec, J. | 2018 | Conference Paper |
| 41 | Graph convolutional networks with argument-aware pooling for event detection | 35 | Nguyen, T.H.; Grishman, R. | 2018 | Conference Paper |
| 42 | A hybrid model for spatiotemporal forecasting of PM 2.5 based on graph convolutional neural network and long short-term memory | 32 | Qi, Y.; Li, Q.; Karimian, H.; Liu, D. | 2019 | Article |
| 43 | Disease prediction using graph convolutional networks: Application to Autism Spectrum Disorder and Alzheimer's disease | 32 | Parisot, S.; Ktena, S.I.; Ferrante, E.; Lee, M.; Guerrero, R.; Glocker, B.; Rueckert, D. | 2018 | Article |
| 44 | Graph-to-sequence learning using gated graph neural networks | 32 | Beck, D.; Haffari, G.; Cohn, T. | 2018 | Conference Paper |
| 45 | Learning to represent programs with graphs | 32 | Allamanis, M.; Brockschmidt, M.; Khademi, M. | 2018 | Conference Paper |
| 46 | MGAE: Marginalized graph autoencoder for graph clustering | 32 | Wang, C.; Pan, S.; Long, G.; Zhu, X.; Jiang, J. | 2017 | Conference Paper |
| 47 | KGAT: Knowledge graph attention network for recommendation | 31 | Wang, X.; He, X.; Cao, Y.; Liu, M.; Chua, T.-S. | 2019 | Conference Paper |
| 48 | Graph neural networks for social recommendation | 31 | Fan, W.; Ma, Y.; Li, Q.; He, Y.; Zhao, E.; Tang, J.; Yin, D. | 2019 | Conference Paper |
| 49 | Deep learning in bioinformatics: Introduction, application, and perspective in the big data era | 30 | Li, Y.; Huang, C.; Ding, L.; Li, Z.; Pan, Y.; Gao, X. | 2019 | Article |
| 50 | DeepInf: Social influence prediction with deep learning | 30 | Qiu, J.; Tang, J.; Ma, H.; Dong, Y.; Wang, K.; Tang, J. | 2018 | Conference Paper |
| 51 | Heterogeneous graph attention network | 29 | Wang, X.; Ji, H.; Cui, P.; Yu, P.; Shi, C.; Wang, B.; Ye, Y. | 2019 | Conference Paper |
| 52 | Distance metric learning using graph convolutional networks: Application toÂ functional brain networks | 29 | Ktena, S.I.; Parisot, S.; Ferrante, E.; Rajchl, M.; Lee, M.; Glocker, B.; Rueckert, D. | 2017 | Conference Paper |
| 53 | Cross-lingual knowledge graph alignment via graph convolutional networks | 28 | Wang, Z.; Lv, Q.; Lan, X.; Zhang, Y. | 2020 | Conference Paper |
| 54 | Spectral graph convolutions for population-based disease prediction | 28 | Parisot, S.; Ktena, S.I.; Ferrante, E.; Lee, M.; | 2017 | Conference Paper |
| | | | Moreno, R.G.; Glocker, B.; Rueckert, D. | | |
| 55 | Graph neural networks for ranking web pages | 28 | Scarselli, F.; Hagenbuchner, M.; Yong, S.L.; Tsoi, A.C.; Gori, M.; Maggini, M. | 2005 | Conference Paper |
| 56 | Situation Recognition with Graph Neural Networks | 27 | Li, R.; Tapaswi, M.; Liao, R.; Jia, J.; Urtasun, R.; Fidler, S. | 2017 | Conference Paper |
| 57 | Point convolutional neural networks by extension operators | 26 | Atzmon, M.; Maron, H.; Lipman, Y. | 2018 | Article |
| 58 | Two-stream adaptive graph convolutional networks for skeleton-based action recognition | 24 | Shi, L.; Zhang, Y.; Cheng, J.; Lu, H. | 2019 | Conference Paper |
| 59 | Convolutional Neural Network Architectures for Signals Supported on Graphs | 24 | Gama, F.; Marques, A.G.; Leus, G.; Ribeiro, A. | 2019 | Article |
| 60 | Compound-protein interaction prediction with end-to-end learning of neural networks for graphs and sequences | 24 | Tsubaki, M.; Tomii, K.; Sese, J. | 2019 | Article |

``